\documentclass[runningheads]{llncs}
\usepackage[T1]{fontenc}
\usepackage{graphicx}
\usepackage{soul}
\usepackage{color}
\usepackage{booktabs}
\usepackage{multicol}
\usepackage{multirow}
\usepackage{amsmath}
\usepackage{ulem}
\usepackage{subcaption}
\usepackage{url}
\begin{document}
\title{CoastTerm: a Corpus for Multidisciplinary Term Extraction in Coastal Scientific Literature}
\titlerunning{CoastTerm}
 \author{Julien DELAUNAY\inst{1,2}\orcidID{0009-0001-9247-5745}$^*$ \and
 Hanh Thi Hong TRAN\inst{1,3,4}\orcidID{0000-0002-5993-1630}$^*$ \and
 Carlos-Emiliano GONZÁLEZ-GALLARDO\inst{1}\orcidID{0000-0002-0787-2990} \and Georgeta BORDEA\inst{1}\orcidID{0000-0001-9921-8234} \and Mathilde DUCOS\inst{1}\orcidID{0009-0003-4355-1531} \and   Nicolas SIDERE\inst{1}\orcidID{0000-0001-6719-5007} \and Antoine DOUCET\inst{1}\orcidID{0000-0001-6160-3356} \and Senja POLLAK\inst{4} \orcidID{0000-0002-4380-0863} \and Olivier DE VIRON \inst{2}\orcidID{0000-0003-3112-9686}}
% %
 \authorrunning{Delaunay et al.}
%\institute{L3i, Univ. La Rochelle, La Rochelle, France \and
 %LIENSs, Univ. La Rochelle, La Rochelle, France \and
 %Jozef Stefan Institute, Ljubljana, Slovenia \\
 %}

 \institute{University of La Rochelle, L3i, La Rochelle, France  \and
University of La Rochelle, LIENSs, La Rochelle, France  \and
Jožef Stefan International Postgraduate School, Ljubljana, Slovenia \and
Jožef Stefan Institute, Ljubljana, Slovenia \\  \email{\{firstname.lastname,thi.tran,carlos.gonzalez\_gallardo\}@univ-lr.fr} \\ \email{senja.pollak@ijs.si}
}

\maketitle              % typeset the header of the contribution
\def\thefootnote{*}\footnotetext{These authors contributed equally to this work}\def\thefootnote{\arabic{footnote}}

\begin{abstract}
%The abstract should briefly summarize the contents of the paper in 150--250 words.
The growing impact of climate change on coastal areas, particularly active but fragile regions, necessitates collaboration among diverse stakeholders and disciplines to formulate effective environmental protection policies. We introduce a novel specialized corpus comprising 2,491 sentences from 410 scientific abstracts concerning coastal areas, for the Automatic Term Extraction (ATE) and Classification (ATC) tasks. Inspired by the ARDI framework, focused on the identification of Actors, Resources, Dynamics and Interactions, we automatically extract domain terms and  their distinct roles  in the functioning of coastal systems by leveraging monolingual and multilingual transformer models. The evaluation demonstrates consistent results, achieving an F1 score of approximately 80\% for automated term extraction and F1 of 70\% for extracting terms and their labels. These findings are promising and signify an initial step towards the development of a specialized Knowledge Base dedicated to coastal areas.

\end{abstract}

\section{Introduction}

% Context
Coastal areas, grappling with the dual impacts of global change and human interventions, constitute a complex system wherein various dynamics continuously interact (physical, chemical, biological, societal, and others). Understanding this system necessitates examining it as an inherently anthropized environment. In this context,  many agents and mechanisms can only be understood by considering human actions, such as coastal development, activities affecting land and sea, resource management, and urbanization.

The resulting interdisciplinarity provides a very interesting use case for the automatic analysis of terminology and its use. Numerous nuanced terms from various domains like environmental science, geography, ecology, and sociology emerge in coastal literature, reflecting its interdisciplinary nature.
Automatic analysis of terminology offers a systematic approach to identify and categorize these domain-specific concepts, crucial to the understanding of the dynamic nature of coastal environments. By adapting to the evolving terminological landscape, this process aids in identifying key entities within coastal systems and integrating diverse disciplinary perspectives into a coherent framework.

% Description of ARDI ontology
The ARDI framework \cite{etienne2011ardi} enables the identification of \textit{Actors} influencing a territory, the \textit{Resources} they exploit, the \textit{Dynamics} in operation, and the \textit{Interactions} between these agents and resources.
Employing ARDI enables stakeholders to collaboratively construct a conceptual framework of the system, facilitating the development of environment protection policies and enhancing scientists' comprehension of these territories. However, identifying key entities within such intricate systems poses considerable challenges. This difficulty is exacerbated by the numerous scientific disciplines addressing the subject matter, including but not limited to biology, oceanography, chemistry, and engineering, resulting in a huge amount of scholarly literature every year.
In this context, automatic term extraction (ATE) and classification (ATC) play a crucial role in unlocking the knowledge embedded within the scientific literature.

Recently, we observed a push towards several ATE approaches with different methods, from rule-based to neural approaches. However, there is still a significant gap in the performance of ATE compared to other similar natural language processing (NLP) downstream tasks partially due to the following reasons. First, terms are inherently semantically defined to refer to domain-specific concepts \cite{kockaert2015handbook}. Besides attempts to define the meaning of a \textit{term}, such as a “language used in a subject field and characterized by the use of specific linguistic means of expression” (ISO 1087-1), in real-life settings, terms are not consistently defined and their definitions vary across the domains and use-cases, making it difficult to develop universal term extraction methods. Secondly, there is a lack of well-documented and transparent domain-specific corpora, even if recently some valuable efforts have been made \cite{rigouts2020termeval}.

% Our constribution

Our main contributions are threefold and are summarized as follows:
\begin{itemize}
    \item We propose two gold-annotated multidisciplinary datasets for ATE and ATC focusing on the domain of coastal areas\footnote{Corpus and code are available at \url{https://github.com/jdelaunay/coastal_area_term_extraction}}.
    \item Inspired by the ARDI framework \cite{etienne2011ardi}, we propose a set of labels designed to facilitate the representation of a system independently of the study domain.
    \item We compare a range of ATE state-of-the-art models on our datasets and identify specific challenges inherent to our data.
\end{itemize}
\section{Related Work}
\subsection{Term extraction datasets}
Several manually annotated monolingual and multilingual domain-specific resources have been developed for term extraction systems \cite{tran2023recent}, notably for scientific domains. 
The ACL Reference Dataset for Terminology Extraction and Classification (ACL RD-TEC) \cite{qasemizadeh2016acl} serves as a benchmark for evaluating term extraction and classification in the scientific literature related to computational linguistics. It includes 300 manually annotated abstracts from articles in the ACL Anthology Reference Corpus (1978-2006), categorized into various classes.
Augenstein et al. (2017) introduce a corpus for SemEval 2017 Task 10, featuring 500 double-annotated documents from the domains of computer science, materials science and physics \cite{augenstein2017semeval}. This corpus addresses the identification and classification of keyphrases at the word level, categorizing keyphrases into \textit{process}, \textit{task}, and \textit{material}.

In relation to environment studies, SPECIES-800 \cite{pafilis2013species} is a corpus of 800 manually annotated abstracts for taxon mentions recognition. Constructed by randomly selecting 100 MEDLINE abstracts from various journals, it comprises 3,708 mentions of 718 unique species, referenced by 1,503 unique names. BiodivNERE \cite{abdelmageed2022biodivnere} offers two gold standard corpora for named entity recognition (NER) and relation extraction (RE) within biodiversity studies, created from biodiversity metadata and abstracts and manually verified by experts. The corpus comprises 2,398 statements from 150 documents, with the NER corpus identifying entities such as \textit{organism}, \textit{environment}, \textit{quality}, \textit{location}, \textit{phenomena}, and \textit{matter}. COPIOUS \cite{nguyen2019copious} stands as a gold standard corpus sourced from the Biodiversity Heritage Library. Comprising more than 26K sentences from 668 documents, it classifies its 28K entities into five categories: \textit{taxon names}, \textit{geographical locations}, \textit{habitats}, \textit{temporal expressions} and \textit{person names}.

However, to our knowledge, no annotated corpora for term extraction in the interdisciplinary study of coastal regions or related domains (such as oceans or seas) exists. Thus, CoastTerm represents a pioneering effort in this regard, paving the way for an exhaustive cross-domain and multi-disciplinary Knowledge Graph construction system for studying coastal areas.

\subsection{Term extraction methods}
% \hl{this section is focused on ATE, but you also mention ATC, is related work in that domain to be coveed as well}
Classical term extractors mainly rely on either linguistic or statistical aspects \cite{frantzi1998c} or combine both \cite{kessler2019extraction} and applied rule-based or machine-learning methods to extract the candidate terms. More recently, the introduction of representation learning and neural networks has led to the application of various text embedding techniques for term extraction, including local-global \cite{amjadian2016local}, non-contextual embeddings (e.g., GloVe\footnote{\url{https://nlp.stanford.edu/projects/glove/}}, Word2Vec, skip-gram \cite{amjadian2016local,amjadian2018distributed,kucza2018term,zhang2018semre}), contextual word embeddings (e.g., Flair\footnote{\url{https://github.com/flairNLP/flair}}, BERT \cite{kucza2018term,andrius2020automatic,terryn2022tagging}), as well as their combinations (e.g., stacked Flair + BERT \cite{andrius2020automatic,terryn2022tagging}).

Neural architectures are also used as end-to-end term extraction systems, and most current systems focus on tagging-based models \cite{hazem2020termeval,lang2021transforming}. In tagging-based mechanisms, the task was formulated as (1) sequence classification where or not the binary label of a term was assigned to each possible n-gram of a fixed length of a given sentence using different variants of BERT-based models (e.g., BERT, RoBERTa, and XLMR); or (2) token classification, where the label was assigned to each word in the given sentence following the IOB annotation format using different language models.

With the advent of transformers, several pre-trained language models have been applied as token classifiers. Above all, XLMR is now considered a benchmark for several languages \cite{tran2022can,trantransformer}. Cross-domain and cross-lingual learning was also applied to these benchmarks to enhance extraction performance in the absence of available annotated data \cite{hazem2022cross,lang2021transforming,tran2022can,tran2022ensembling,tran2024can,trantransformer}.

In addition, there have been other experiments on ATE with the adoption of span-based methods \cite{wang2023extract} or applying generative models considering Seq2Seq models such as mBART \cite{lang2021transforming} to extract candidate terms more efficiently. However, while span-based methods show their potential, the performance of generative models is still under question.

% Domain specific
In the context of our study domains, Zhao et al. (2022) explores the extraction of knowledge from operational maritime decision-making sentences \cite{10.1007/978-981-19-6052-9_75}, Andersen et al. (2022) proposes the development of a corpus to cultivate a specialized terminology relevant to Norwegian maritime discourse \cite{andersen2022utilising}, Mouratidis et al. (2022) performs term extraction within legal documents on maritime topics in the Greek language \cite{10.1145/3549737.3549751}. Similarly to our work, but in the realm of karst studies, TermFrame\footnote{\url{https://termframe.ff.uni-lj.si/}} serves as a KG constructed by extracting terms and triplets from English and Slovene karst corporas \cite{pollak2019karst,vintar2022framing}. EcoLexicon \cite{faber2013representing,faber2016ecolexicon} constitutes a KB specialized on environment, encompassing six languages (English, French, German, Modern Greek, Russian, and Spanish).
Moreover, for NER tasks, TaxoNERD \cite{le2022taxonerd} aims to recognize taxon mentions in ecological documents, while AGRONER \cite{veena2023agroner} employs unsupervised NER techniques tailored to the agriculture domain, integrating an extended BERT model with Latent Dirichlet Allocation (LDA) topic modeling.
\section{CoastTerm corpus for term extraction}

\subsection{Annotation process}
We collected a corpus of 64,000 papers from Scopus\footnote{\url{https://www.elsevier.com/fr-fr/solutions/scopus}}, spanning from 1980 to 2023, containing the terms ``coastal areas'' or ``littoral'' in their abstract or title.
We initially selected randomly 600 abstracts for manual annotation, keeping a proportion of 60\% regular articles and 40\% surveys.
Annotation efforts first enlisted the participation of two undergraduate Master's students specialized in Earth Sciences, who were compensated for their contributions. These students were supplied with example annotations and guidelines to conduct a task of document-level joint entity and relation extraction, which involves nested ATE and ATC, coreference resolution, and document-level relation extraction. The annotation guidelines were constructed by a computer science and coastal research PhD student and a domain expert, with the assistance of an ontology expert. Only the sentences that give information about the functioning of the coastal zone were annotated, we avoided sentences that described the methodology used in the article. %Two versions of the guidelines were provided, a long version with figures and examples and a short one with only the guidelines and a few examples.

We adapted the ARDI framework to extract the information related to the functioning of a described system within a scientific abstract.
As a result, for ATC, we designated the following labels: ``Actor'' (stakeholders who consume resources and/or initiate processes) and ``Resource'' (goods, products, facilities, and elements, including plants and animals, utilized by stakeholders). Following the Basic Formal Ontology (BFO) \cite{arp2015building}, we replaced the ``Dynamic'' label in ARDI with ``Process'', as defined by the Environment Ontology (ENVO) \cite{buttigieg2013environment}, referring to environmental, societal, or economic processes impacting the system and inducing changes. To enhance precision on the extracted information, we introduced the ``Quality'' label from the Phenotype And Trait Ontology (PATO) \cite{10.1093/bib/bbx035} , which refers to height, concentration, or a specificity, and added a ``Location'' label.

Along with the two students specializing in Earth Sciences, the annotation process engaged the same domain expert and PhD student, all simultaneously annotating papers. All annotators were familiar with the annotation tool, INCEpTION \cite{klie-etal-2018-inception}. Over two months, the campaign aimed to achieve dual-annotator coverage for 60\% of the total annotated abstracts. Ultimately, 215 abstracts were annotated, with a mean Krippendorff's alpha \cite{krippendorff2011computing} of 43\%, indicating a moderate agreement, but not sufficient to use directly the dataset, falling short of the minimum threshold (66\%) required for direct dataset utilization, highlighting the difficulty of the manual annotation of terms. Annotations were then curated by the PhD student according to the guidelines.

We employed three domain-relevant knowledge bases (KBs)  (i.e., AGROVOC\footnote{\url{https://agrovoc.fao.org/browse/agrovoc/en/}}, GEMET (GEneral Multilingual Environmental Thesaurus)\footnote{\url{https://www.eionet.europa.eu/gemet/en/about/}}, AFO (Agriculture and Forestry Ontology)\footnote{\url{https://finto.fi/afo/en/}}, and TAXREF-LD\footnote{\url{https://github.com/frmichel/taxref-ld}}) to pre-annotate pertinent terms in a secondary subset of 195 abstracts. This process significantly assisted in refining term boundaries. Annotations were then carried out by the PhD student annotator following the established guidelines, utilizing insights acquired from the initial annotation process. Subsequently, the initially fully manually annotated subset was homogenized with the KB-recommended one to produce two datasets intended for the study of coastal areas. In the context of the present research, both datasets were adapted manually for terminology extraction by removing relations and pronouns that indicate coreferences.

\subsection{Dataset description}

\begin{table}[t]
\centering
\caption{Statistics for KB and human recommended subcorporas}
\label{table::corpus_stats}
\scalebox{0.8}{
\begin{tabular}{{l|c|c}}
\toprule
\midrule
 & \begin{tabular}[c]{@{}l@{}}\textbf{KBs} \textbf{recommended}\end{tabular} & \begin{tabular}[c]{@{}l@{}}\textbf{Human experts}  \textbf{recommended}\end{tabular}  \\ 
\midrule
Vocabulary size           & 7,902 & 6,280 \\
Sentences                 & 1,235 & 1,256 \\ 
Tokens                    & 31,147 & 37,983 \\ 
Type-to-token ratio (TTR) & 0.25 & 0.17\\ 
Annotated Terms           & 6,663 & 6,543 \\ 
Unique Terms              & 3,844 & 4,400\\
Unique Terms (lemmatized) & 3,539 & 4,110\\
\# of ``Actor'' & 617 & 602 \\
\# of ``Resource'' & 2,236 & 2,052 \\ 
\# of ``Process'' & 1,524 & 1,431 \\ 
\# of ``Quality'' & 1,138 & 1,373 \\
\# of ``Location'' & 1,145 & 1,082 \\
\# of ``B'' & 6,663 & 6,543  \\
\# of ``I'' & 4,723 & 6,192 \\
\# of ``O'' & 25,415 & 25,710 \\
\midrule
\bottomrule
\end{tabular}
}
\end{table}

\begin{figure}
    \centering
    \includegraphics[scale=0.33]{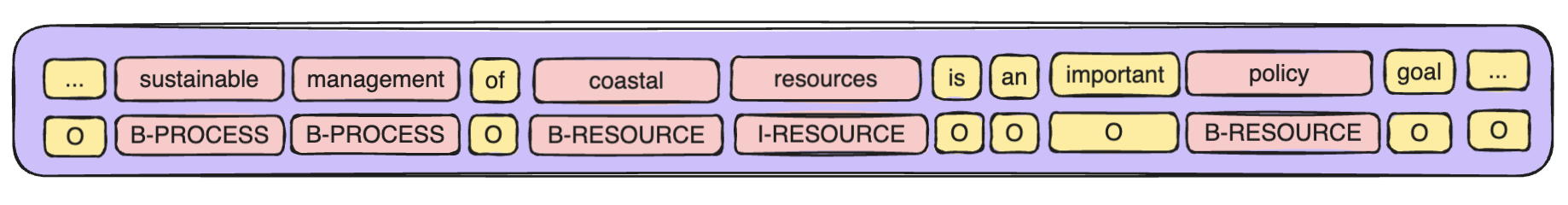}
    \caption{A sample of the corpus annotation for term extraction and classification}
    \label{fig:iob_example}
\end{figure}

The KB-recommended corpus contains 1,235 annotated sentences, spanning across 61 different keywords, and includes 6,663 annotated terms. The human-recommended corpus contains 1,256 annotated sentences, spanning across 92 different keywords, and featuring 6,543 annotated terms. From Table \ref{table::corpus_stats} it can be seen that the distribution of labels is consistent across both datasets. In addition, the label ``Actor'' is observed to be less prevalent, whereas ``Resource'' emerges as the most frequently represented one. 
Together, these two datasets collectively span across 101 unique keywords, ``\textit{aquatic science}'', ``\textit{oceanography}'', ``\textit{ecology, evolution, behavior, and systematics}'', ``\textit{ecology}'', ``\textit{pollution}'', ``\textit{earth and planetary sciences}'', ``\textit{environmental science}'', ``\textit{management, monitoring, policy, and law}'', ``\textit{water science and technology}'', and ``\textit{geography, planning, and development}'' being the most prominent.
A comparison of the top 100 frequent terms shows a significant overlap of 52\% between the two corpora.
In total, the two datasets share 751 common terms (730 if lemmatized).
We allocated 70\% of the annotated sentences within each article to the training sets, 10\% to the validation sets, and 20\% to the test sets. An annotation example can be visualized in Figure \ref{fig:iob_example}.

%\input{tables/iob}

% \begin{figure}[ht!]
%     \centering
%     \begin{subfigure}{0.49\textwidth}
%         \centering
%         \includegraphics[width=\textwidth]{img/kb_wordcloud.png}
%         \subcaption{KB}
%     \end{subfigure}
%     \hfill
%     \begin{subfigure}{0.49\textwidth}
%         \centering
%         \includegraphics[width=\textwidth]{img/human_wordcloud.png}
%         \subcaption{Human}
%     \end{subfigure}
%     \caption{Wordcloud of the $n=100$ most frequent lemmatized terms.}
%     \label{fig:pos_wordcloud}
% \end{figure}

\vspace{-20pt}

\section{Experiments}
To evaluate the challenges of CoastTerm, we conducted extensive experiments using state-of-the-art ATE, extending our evaluation to ATC.
Our in-depth analysis facilitates a discussion on potential future trajectories for interdisciplinary ATE and ATC research related to coastal areas.

\subsection{Models}

We considered ATE as a sequence-labeling task where the model returns a label for each token in a text sequence, using the IOB labeling mechanism \cite{rigouts2021hamlet,tran2022can,tran2022ensembling}. % where B stands for the beginning word in the term, I stands for the word inside the term, and O stands for the word not part of the term. 
We apply the same labeling scheme to ATC, with the addition of the term's class following its IOB label.

% The terms from a gold standard list are first mapped to the tokens in the raw text and each word inside the text sequence is annotated with one of the three labels. In general, the model is first trained to predict a label for each token in the input text sequence (e.g., we model the task as token classification) and then applied to the unseen text (test data). Finally, from the tokens or token sequences labeled as terms, the final candidate term list for the test data is composed. 

We experimented with two families of language models with both base and large versions, including:

\begin{itemize}
    \item \textit{Monolingual pre-trained model}: We chose RoBERTa \cite{liu2019roberta}, a transformer-based model pre-trained on a large corpus of English data in a self-supervised fashion. %This means that it was pretrained only on the raw texts only (no human labeling needed) with an automatic process to generate inputs and labels from those texts.
    \item \textit{Multilingual pre-trained model}: We opt for XLMR \cite{conneau2019unsupervised}, a transformer-based model pre-trained on 2.5TB of filtered CommonCrawl data containing 100 languages. This multilingual version of RoBERTa, achieves benchmark performance in ATE for rich-resourced languages (e.g. English) \cite{rigouts2020termeval,tran2022can}.
%    \item \textit{Large-scale pre-trained model}: LS-Llama\footnote{https://github.com/4AI/LS-LLaMA}
\end{itemize}

\subsection{Evaluation Metrics}
To assess the performance of the ATE systems on the human-recommended and KB-recommended datasets, we juxtaposed the candidate list of unique terms extracted from the entire test set against the gold standard of the test set. We performed experiments with both lemmatized and not lemmatized terms for testing. This evaluation was conducted employing strict matching criteria, with metrics including precision (P), recall (R), and micro F1-score (F1). For the ATC task, the evaluation process remained the same but each term was accompanied by its corresponding label.

% We recall the metrics formulae:

% $\text{Precision} = \frac{\text{True Positives}}{\text{True Positives + False Positives}}$; $\text{Recall} = \frac{\text{True Positives}}{\text{True Positives + False Negatives}}
% $

% \vspace{10pt}
% $\text{F1-score} = 2 \times \frac{\text{Precision} \times \text{Recall}}{\text{Precision + Recall}}
% $

\subsection{Results}

% ATE
Table \ref{tab:results} presents the performances of mono- and multilingual classifiers for the ATE task on both the human- and KB-recommended datasets. Regarding the KB-recommended dataset, the results demonstrate that XLMR$_{base}$ achieves superior performance compared to RoBERTa$_{base}$ for both lemmatized and unlemmatized text. However, RoBERTa$_{large}$ outperforms XLMR$_{large}$. Conversely, on the human-recommended dataset, monolingual models exhibit better performance than multilingual ones. Additionally, it is observed that XLMR$_{large}$ performs least effectively in this context.
We also performed experiments with the combined KB- and human-recommended datasets. This configuration is referred to as Fusion in Table \ref{tab:results} where 20\% and 40\% of the human-recommended dataset were allocated to the validation and test sets respectively. The KB-recommended dataset, along with the remaining 40\% of the human-recommended dataset, composed the train set. Reported performances show that monolingual models outperform multilingual ones.

% ATC
Table \ref{tab:results_classif} displays the performance of the classifiers for the ATC task using the combined KB-recommended and human-recommended datasets. Notably, multilingual classifiers demonstrate superior performance compared to monolingual ones. Additionally, upon lemmatizing the predictions, XLMR$_{base}$ exhibits better performance than XLMR$_{large}$.
\vspace{-10pt}

\begin{table}[t]
\centering
\caption{Evaluation in performance of different extractors for ATE task}
\label{tab:results}
\scalebox{0.75}{
\begin{tabular}{l|lll|lll|lll}
\toprule
    % & \multicolumn{9}{c|}{\bf Not lemmatized} & \multicolumn{9}{c}{\bf Lemmatized} \\
\midrule
\multirow{2}{*}{\bf Models}  & \multicolumn{3}{c|}{\bf KB}                 & \multicolumn{3}{c|}{\bf Human} & \multicolumn{3}{c}{\bf Fusion} \\ 
                    & \bf P & \bf R & \multicolumn{1}{l|}{\bf F1} & \bf P & \bf R &  \multicolumn{1}{l|}{\bf F1} & \bf P    & \bf R    & \bf F1   \\ 
\midrule
 \multicolumn{10}{c}{\bf Not lemmatized} \\
\midrule
RoBERTa$_{base}$   & 75.51 & 78.12 &   76.79 & 78.65 & 82.07 & 80.32 & \bf 79.08 & 80.37 & 79.72 \\
XLMR$_{base}$  &  76.08 & 78.95 &  \multicolumn{1}{l|}{77.49} & 79.07 & 81.04 & 80.04 & 77.95 & 79.94 & 78.93   \\
RoBERTa$_{large}$  & \bf 77.23 & \bf 79.90 & \bf 78.54 & \bf 80.97 & \bf 82.90 & \bf 81.92 & 79.00 & 80.96 & \bf 79.97  \\
XLMR$_{large}$  & 76.99 & 78.85 & 77.91 & 77.42 & 79.59 & 78.49  & 78.48 & \bf 81.45 & 79.94 \\
\midrule
\multicolumn{10}{c}{\bf Lemmatized} \\
\midrule
RoBERTa$_{base}$  & 75.45 & 78.24 & \multicolumn{1}{l|}{76.82} & 78.43 & 82.25 & \multicolumn{1}{l|}{80.29} & \bf 78.80 & 79.97 & 79.38 \\
XLMR$_{base}$    & 75.89 & 78.78 & \multicolumn{1}{l|}{77.31} & 78.97 & 81.28 & \multicolumn{1}{l|}{80.11} & 77.61 & 79.68 & 78.63 \\
RoBERTa$_{large}$  &  \bf 77.32 & \bf 79.76 & \multicolumn{1}{l|}{\bf 78.52} & \bf 80.78 & \bf 82.79 & \multicolumn{1}{l|}{\bf 81.77} & 78.74 & 80.70 & \bf 79.71 \\
XLMR$_{large}$  &  76.93 & 79.11 &  \multicolumn{1}{l|}{78.00} & 77.48 & 80.41 & \multicolumn{1}{l|}{78.92} & 78.20 & \bf 81.21 & 79.68  \\
\midrule
\bottomrule
\end{tabular}
}
\end{table}

%\vspace{-20pt}
\begin{table}[ht]
\centering
\caption{Evaluation in performance of different extractors for joint ATE and ATC on the fusion of KB and human datasets}
\label{tab:results_classif}
\scalebox{0.8}{
\begin{tabular}{l|lll|lll}
\toprule
\midrule
    & \multicolumn{3}{c|}{\bf Not lemmatized} & \multicolumn{3}{c}{\bf Lemmatized} \\
% \midrule
\bf Models  & \bf P & \bf R & \multicolumn{1}{l|}{\bf F1} & \bf P    & \bf R    & \bf F1 \\ 
\midrule
RoBERTa$_{base}$   &  66.36 & 68.93 & \multicolumn{1}{l|}{67.62} & 65.96 & 68.74 & 67.32 \\
XLMR$_{base}$   & 66.11 & \bf 71.69 & \multicolumn{1}{l|}{68.79} & 65.95 & \bf 71.85 & \bf 68.77  \\
RoBERTa$_{large}$  &  66.60 & 71.05 & \multicolumn{1}{l|}{68.75} & 66.05 & 70.85 & 68.37  \\
XLMR$_{large}$  & \bf 66.97 & 71.16 & \multicolumn{1}{l|}{\bf 69.00} & \bf 66.60 & 70.96 & 68.71  \\
\midrule
\bottomrule
\end{tabular}
}
\end{table}
\vspace{-20pt}

\subsection{Error analysis}

\begin{figure}[ht!]
    \centering
    \includegraphics[scale=0.3]{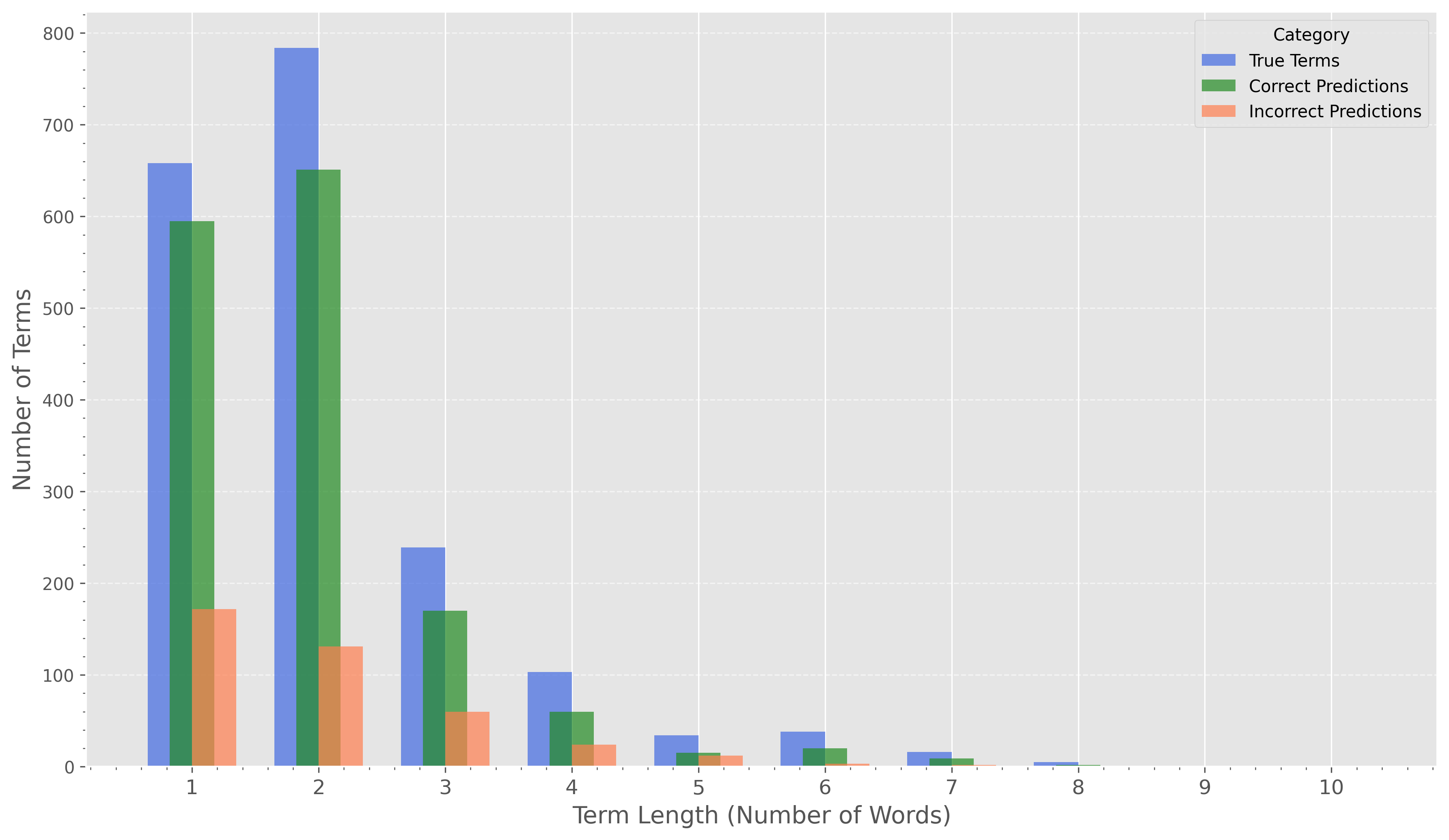}
    \caption{Term length distribution for XLMR$_{base}$ on the fusion of KB- and human-recommended datasets}
    \label{fig:term_length}
\end{figure}

To assess the impact of term length on the models' performance, we examined the proportion of correct and incorrect predictions relative to term length. Figure \ref{fig:term_length} illustrates for XLMR$_{base}$ that as the term length increases, the model encounters greater difficulty in accurately predicting it and might predict shorter terms. However, the vast majority of the terms consist of either one or two words which mitigates the negative impact on performance. This behavior is observed across all models.
%\vspace{-20pt}
%\vspace{-40pt}
\begin{figure}[h]
    \centering
    \begin{subfigure}{0.5\textwidth}
        \centering
        \includegraphics[width=\linewidth]{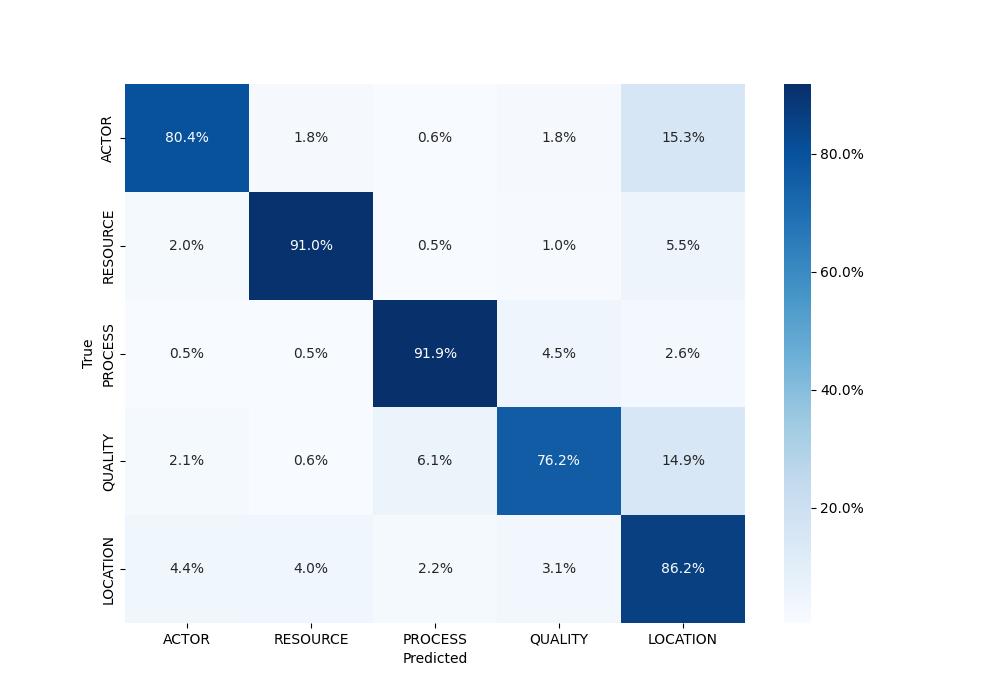}
        \caption{Normalized on ground truth}
        \label{fig:sub1}
    \end{subfigure}%
    \begin{subfigure}{0.5\textwidth}
        \centering
        \includegraphics[width=\linewidth]{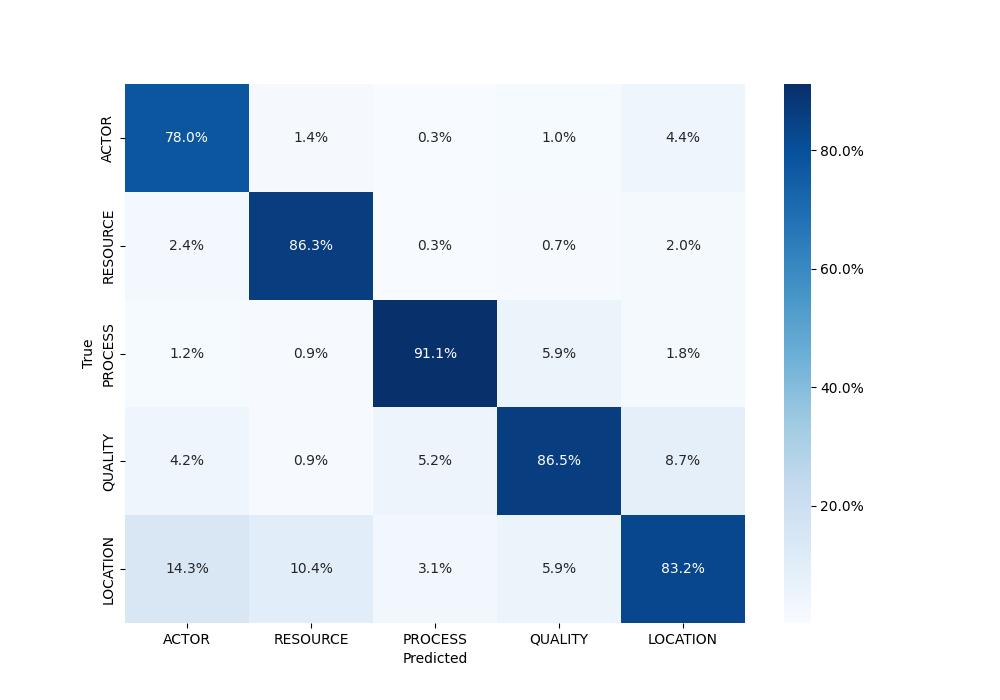}
        \caption{Normalized on predictions}
        \label{fig:sub2}
    \end{subfigure}
    \caption{Confusion matrices for XLMR$_{base}$ on the fusion of KB- and human-recommended datasets}
    \label{fig:confusion_matrix}
\end{figure}
%\vspace{-20pt}

We also report the confusion matrices for the ATC in Figure \ref{fig:confusion_matrix}. The results are consistent since ambiguity might come from the fact that one entity might be an ``Actor'' at one point and a ``Location'' at another (Countries, for example); or a ``Location'' and a ``Resource'' or ``Quality'' depending on the context.

\section{Conclusion and Future Works}

We introduced CoastTerm, a corpus of 2,491 gold-annotated sentences for interdisciplinary automatic term extraction and classification related to the coastal area. 
%The corpus will be made available via ZENODO\footnote{https://zenodo.org/} with the non-anonymous camera ready version. 
Adapting the ARDI framework, we provided comprehensive labels applicable to various domains. Benchmarking mono- and multilingual state-of-the-art models on ATE and ATC tasks shows promising results, paving the way for interdisciplinary knowledge base construction in the coastal area domain.

% \section{Acknowledgments}
% Acknowledgements will be added for non-anonymous version.
\subsection*{Acknowledgments}
The work was supported by the TERMITRAD (2020-2019-8510010) project funded by the Nouvelle-Aquitaine Region, France, by the Slovenian Research and Innovation Agency core research program Knowledge Technologies (P2-0103) and the project Cross-lingual and Cross-domain Methods for Terminology Extraction and Alignment, a bilateral project funded by the program PROTEUS under the grant number BI-FR/23-24-PROTEUS006. We express our gratitude to Kenza HERMAN and Geraldine DUBOS for their invaluable assistance in the annotation process.

\bibliographystyle{splncs04}
\bibliography{main}
\end{document}